\documentclass[10pt, journal, final]{IEEEtran}
\usepackage{cite}
\usepackage{graphicx}
\usepackage{amsmath,amsfonts,amssymb,amsthm,amsbsy}
\usepackage{color}
\usepackage{enumerate}
\usepackage{subfigure}
\usepackage{multirow}
\usepackage{algorithmic, algorithm}

\theoremstyle{plain}

\theoremstyle{definition}

\theoremstyle{remark}
\newcounter{REM}
\newtheorem{rem}[REM]{Remark}

\newcommand{\btheta}{\boldsymbol{\theta}}
\newcommand{\bfeta}{\boldsymbol{\eta}}
\newcommand{\bmu}{\boldsymbol{\mu}}
\newcommand{\bxi}{\boldsymbol{\xi}}

\begin{document}
\title{\huge{Bayesian Learning Approach to Model Predictive Control}}

\author{Namhoon Cho, Seokwon Lee, Hyo-Sang Shin, and Antonios Tsourdos
	\thanks{Namhoon Cho, Seokwon Lee, Hyo-Sang Shin, and Antonios Tsourdos are with the Centre for Autonomous and Cyber-Physical Systems, School of Aerospace, Transport and Manufacturing, Cranfield University, Cranfield, MK43 0AL, Bedfordshire, United Kingdom. e-mail: \{n.cho, seokwon.lee, h.shin, a.tsourdos\}@cranfield.ac.uk}
}

\maketitle

\begin{abstract}
	This study presents a Bayesian learning perspective towards model predictive control algorithms. High-level frameworks have been developed separately in the earlier studies on Bayesian learning and sampling-based model predictive control. On one hand, the Bayesian learning rule provides a general framework capable of generating various machine learning algorithms as special instances. On the other hand, the dynamic mirror descent model predictive control framework is capable of diversifying sample-rollout-based control algorithms. However, connections between the two frameworks have still not been fully appreciated in the context of stochastic optimal control. This study combines the Bayesian learning rule point of view into the model predictive control setting by taking inspirations from the view of understanding model predictive controller as an online learner. The selection of posterior class and natural gradient approximation for the variational formulation governs diversification of model predictive control algorithms in the Bayesian learning approach to model predictive control. This alternative viewpoint complements the dynamic mirror descent framework through streamlining the explanation of design choices.
\end{abstract}

\begin{IEEEkeywords}
	Bayesian learning, variational inference, posterior approximation, natural gradient descent, mirror descent, model predictive control, online learning, sampling
\end{IEEEkeywords}

\section{Introduction} \label{Sec:Intro}
A Bayesian perspective towards machine learning is to regard Bayesian methods as the solution to an optimisation problem associated with the information geometry of the posterior. A general way of understanding the Bayesian principle through the reformulated sequential optimisation-oriented view was originally populated in \cite{Jaynes_1982,Jaynes_1988} as the principle of maximum-entropy. Noticing from such fundamental principle, the work in \cite{Khan_2021} showed that many machine learning methods being used have a Bayesian nature. This gives rise to the known benefits in robustness and flexibility of the learning algorithms in the real world where the information is presented not all at once while the world keeps changing. The Bayesian learning rule (BLR) is then presented as an unifying framework that can be applied ubiquitously to the optimisation problems with a recurring pattern so that it defines a family of machine learning algorithms. In this framework, the algorithms derived in non-Bayesian settings are understood as the special cases where the temperature parameter is set to zero so that the entropy term in the cost function vanishes. What is more important to the context of this study is that \cite{Khan_2021} noted about the connection of the BLR to online learning.

Meanwhile, the work of \cite{Wagener_2019} showed that model predictive control (MPC) can be viewed from an online learning perspective where the agent makes an action in response to the loss returned from the environment as a cumulative result of previous actions. With this view, many existing MPC algorithms are interpreted as special instances of applying dynamic mirror descent (DMD) method to the associated optimisation problem. The DMD-MPC framework is thus capable of generating a wide range of MPC algorithms, most notably, including the model predictive path integral (MPPI) control algorithm \cite{Williams_2016,Williams_2017a,Gandhi_2021}, the information-theoretic MPC algorithms \cite{Williams_2017b,Williams_2018}, and the cross-entropy method. The DMD-MPC algorithm itself is incorporated as the inner-loop policy with a model-free reinforcement learning outer-loop policy in \cite{Mishra_2021}. More broadly, MPC is related to online rollout on top of offline training \cite{Bertsekas_2021}. Other recent studies have also investigated MPC in the perspective of online learning where the learner updates itself in response to the loss signal given by the environment and analysed the regret bounds \cite{Yu_2020,Shi_2020,Lin_2021,Shi_2021}.

With this background, it has now been clear that both learning and control are centred around some key principles in common; i) formulation of a stochastic optimisation problem, and ii) employment of a Bayesian inference method to solve the posed problem. This work aims to strengthen this unified view by combining the insights gained from the BLR and the DMD-MPC to establish a Bayesian learning approach to control problems in MPC setup with continuous-time formulation. This study presents the Bayesian learning MPC (BL-MPC) as a generic theoretical framework for generating sample-rollout-based MPC algorithms. Notably, design diversification in the BL-MPC arises quite naturally from the choice of posterior and natural gradient approximation for variational inference, whereas the DMD-MPC separates the choice of control sampling distribution class and the Bregman divergence term in the objective function. The BL-MPC approach is thus a complementary culmination of the connections between Bayesian learning, online learning, and MPC.

The rest of the paper is organised as follows: Section \ref{Sec:ProbForm} briefly presents the problem formulation, and Sec. \ref{Sec:BL-MPC} presents the Bayesian learning model predictive control framework. The Bayesian approach is discussed in relation to the BLR and the DMD-MPC. To demonstrate how the BL-MPC approach can generate different algorithms, Sec. \ref{Sec:GaussExample} provides a concrete example considering Gaussian distribution for posterior approximation. Section \ref{Sec:Concls} concludes the paper.

\section{Problem Formulation} \label{Sec:ProbForm}
For the development of the BL-MPC framework, let us first consider the following optimal control problem
\begin{equation} \label{Eq:ProbForm_DeterSys}
	\begin{aligned}
		&\text{minimise}	&	J\left(\btheta\right) &= \mathbb{E}_{p_{u}}\left[C_{t:t_{f}}\left(\btheta\right) + R\left(\btheta\right)\right]\\
		&\text{subject to} &	\dot{\mathbf{x}}\left(\tau\right) &= \mathbf{f}\left(\tau,\mathbf{x}\left(\tau\right),\mathbf{u}\left(\tau\right)\right)\\
		&	&	\mathbf{x}\left(t\right) &= \mathbf{x}_{t}\\
		&	&	\mathbf{u}\left(\tau\right) &\sim p_{u}\left( \btheta | \left[t,t_{f}\right]\right) 
	\end{aligned}
\end{equation}
where $\tau$ denotes the time variable that evolves as the independent variable, $t$ denotes the current time at which the planning operation and the control execution takes place, $t_{f} = t + T$ denotes the final time in the planning horizon of length $T$, $\mathbf{x}\left(t\right) \in \mathbb{R}^{n \times 1}$ denotes the vector-valued state function, $\mathbf{x}_{t}$ denotes the state value at $t$, $\mathbf{u}\left(t\right) \in \mathbb{R}^{m \times 1}$ denotes the vector-valued control input function or its value at $t$, and $\btheta$ represents the random variable introduced in the policy. In the above formulation, $J\left(\btheta\right)$ represents the objective function, $C_{t:t_{f}}\left(\btheta\right)$ represents the cost evaluated for each state trajectory spanning the time interval $\left[t,t_{f}\right]$, and $R\left(\btheta\right)$ represents the regulariser. The trajectory cost is usually defined in the Bolza form passed through a utility function as 
\begin{equation} \label{Eq:TrajCost}
	\resizebox{0.91\hsize}{!}{$\displaystyle
	C_{t:t_{f}}\left(\btheta\right) := -U\left[\phi\left(\mathbf{x}\left(t_{f}\right)\right) + \int_{t}^{t_{f}} L\left(\tau, \mathbf{x}\left(\tau\right),\mathbf{u}\left(\tau\right)\right)d\tau\right]
	$}
\end{equation}
with some convex functions $\phi$, $L$, and the utility function $U$ defined either simply as i) $U\left(\mathcal{C}\right) = -\mathcal{C}$ for $\forall \mathcal{C} \geq 0$ or as ii) a monotonically decreasing function satisfying
\begin{equation} \label{Eq:utility}
	\begin{aligned}
		U: \mathbb{R}_{+} &\mapsto \left[0,1\right]\\
		U\left(0\right) &= 1\\
		\lim\limits_{\mathcal{C}\rightarrow\infty} U\left(\mathcal{C}\right) &= 0
	\end{aligned}
\end{equation}

The above formulation considers a deterministic system dynamics $\mathbf{f}$ with a stochastic policy function $p_{u}\left( \btheta | \left[t,t_{f}\right]\right)$ which is described as a parametric probabilistic distribution over functions $\mathbf{u}\left(\tau\right)$ (rather than pointwise evaluations of function) in the closed interval $\tau \in \left[t,t_{f}\right]$. The current time instance $t$ in Eq. \eqref{Eq:ProbForm_DeterSys} is the initial time of the planning window, and $\mathbf{x}_{t}$ is considered the given initial state under the assumption that the full state vector is measurable in real-time.

\begin{rem}
	For the purpose of developing a framework with conceptual clarity, this study considers deterministic dynamics with stochastic policy at this stage. Nevertheless, the case of the system dynamics being a stochastic transition function can be addressed in a similar manner by including uncertainties $\mathbf{w}\left(t\right)$ entering into the state transition dynamics as
	\begin{equation} \label{Eq:ProbForm_StochSys}
		\begin{aligned}
		\dot{\mathbf{x}}\left(\tau\right) &= \mathbf{f}\left(\tau,\mathbf{x}\left(\tau\right),\mathbf{u}\left(\tau\right),\mathbf{w}\left(\tau\right)\right)\\
		\mathbf{w}\left(\tau\right) &\sim p_{w}\left(\left[t,t_{f}\right]\right)
		\end{aligned}
	\end{equation}
	and then by taking expectation operation in the definition of cost not only with respect to the control distribution but also with respect to the distribution of uncertain disturbances.
\end{rem}

The MPC setup for the optimal control problem is to rapidly update a simple policy at each time instead of finding a fixed state-dependent policy that performs well over a wide range of operating conditions. An algorithm for updating the policy parameter is needed if the policy $p_{u}$ is defined to be a parametric distribution. Let us use the hat notation $\hat{\left(\right)}$ to denote a predicted or an estimated object. Given a control function $\hat{\mathbf{u}}\left(\tau\right)$ that follows $p_{u}\left(\btheta | \left[t,t_{f}\right]\right)$, the corresponding state trajectory can be predicted by simulating the estimated model for the system dynamics $\hat{\mathbf{f}}$ forward in time from the initial value $\mathbf{x}_{t}$, and it can be expressed as the solution to the following integral equation.
\begin{equation} \label{Eq:PredTraj}
	\hat{\mathbf{x}}\left(\tau\right) = \mathbf{x}_{t} + \int_{t}^{\tau} \hat{\mathbf{f}}\left(\xi, \hat{\mathbf{x}}\left(\xi\right), \hat{\mathbf{u}}\left(\xi\right) \right)d\xi
\end{equation}

\section{Bayesian Learning Model Predictive Control} \label{Sec:BL-MPC}
This section presents the BL-MPC approach by taking inspirations from the flexibility of the BLR as an algorithm generator. \cite{Khan_2021} argues that the BLR is a single mathematical rule which can derive many optimisation / machine learning algorithms as special cases of the BLR. The BLR framework is described as a two-stage scheme where i) an objective function defined in a Bayesian sense is optimised to find posterior approximation and ii) the natural gradient descent is used for optimisation. In this sense, a user or an automated agent should choose the form of i) the posterior approximation and ii) the natural gradient approximation in order to realise the BLR. Thus, taking different choices for the posterior and the natural gradient approximations leads to different optimisation algorithms. 

\subsection{Problem Reformulation}
\subsubsection{Bayesian Objective}
In the Bayesian approach to learning problems, the main objective is to find the posterior distribution $p\left(\btheta | \mathcal{D}\right)$ where $\btheta$ is the random variable representing uncertain quantities that need to be estimated, or to be optimised in some contexts, and $\mathcal{D}$ is the dataset that carries information related to infer, or to find, $\btheta$. 

The posterior distribution computation can be performed in principle with the Bayes rule which is a general mathematical concept that follows from the definition of conditional distribution. The Bayesian update can be generally stated as
\begin{equation} \label{Eq:Bayes_original}
	p\left(\btheta | \mathcal{D}\right) = \frac{p\left(\mathcal{D} | \btheta\right) p\left(\btheta\right)}{p\left(\mathcal{D}\right)} =\frac{p\left(\mathcal{D} | \btheta\right) p\left(\btheta\right)}{\int p\left(\mathcal{D} | \btheta\right) p\left(\btheta\right) d\btheta}
\end{equation}
where $p\left(\btheta\right)$ is the prior distribution for the parameter, $p\left(\mathcal{D} | \btheta \right)$ is the likelihood, $p\left(\mathcal{D}\right)$ is the model evidence also known as the marginal likelihood. 

As it was acknowledged in the study on the BLR \cite{Khan_2021}, the maximum-entropy principle established in \cite{Jaynes_1982,Jaynes_1988} suggests another equivalent approach to the same posterior distribution computation problem by leveraging reformulation into an optimisation problem defined over the set $\mathcal{P}$ of entire probability distributions. First, let a probability distribution $q\left(\btheta\right)$ belongs to $\mathcal{P}$. By definition, the Kullback-Leibler (KL) divergence defined as the relative entropy from the posterior distribution $p\left(\btheta | \mathcal{D}\right)$ to $q\left(\btheta\right)$ is given by
\begin{equation} \label{Eq:KL_div}
	\mathbb{D}_{KL}\left[ \left. q\left(\btheta\right) \right\| p\left(\btheta | \mathcal{D}\right)\right] = \mathbb{E}_{q\left(\btheta\right)}\left[\log \frac{q\left(\btheta\right)}{p\left(\btheta | \mathcal{D}\right)}\right]
\end{equation}
The probability distribution $q^{*}\left(\btheta\right)\in\mathcal{P}$ minimising the KL divergence in Eq. \eqref{Eq:KL_div} is $q^{*}\left(\btheta\right)=p\left(\btheta | \mathcal{D}\right)$. Rearranging Eq. \eqref{Eq:KL_div} by substituting Eq. \eqref{Eq:Bayes_original}, we have
\begin{equation} \label{Eq:KL_div_rearrange}
	\begin{aligned}
	&\mathbb{D}_{KL}\left[ \left. q\left(\btheta\right) \right\| p\left(\btheta | \mathcal{D}\right)\right] = \mathbb{E}_{q\left(\btheta\right)}\left[\log \frac{q\left(\btheta\right)}{\frac{p\left(\mathcal{D} | \btheta\right) p\left(\btheta\right)}{p\left(\mathcal{D}\right)}}\right]\\
	&= - \mathbb{E}_{q\left(\btheta\right)}\left[\log p\left(\mathcal{D} | \btheta\right)\right] + \mathbb{E}_{q\left(\btheta\right)}\left[\log \frac{q\left(\btheta\right)}{p\left(\btheta\right)}\right]\\
	&\quad + \mathbb{E}_{q\left(\btheta\right)}\left[\log p\left(\mathcal{D}\right)\right]\\
	&= - \mathbb{E}_{q\left(\btheta\right)}\left[\log p\left(\mathcal{D} | \btheta\right)\right] + \mathbb{D}_{KL}\left[\left. q\left(\btheta\right) \right\| p\left(\btheta \right)\right]\\
	&\quad + \log p\left(\mathcal{D}\right)
	\end{aligned}
\end{equation}
Since the last term $\log p\left(\mathcal{D}\right)$ is a constant, the posterior distribution can be stated as the optimiser to the following minimisation problem
\begin{equation} \label{Eq:Bayes_reform_objective}
	\begin{aligned}
	J_{B}\left(\btheta\right) &:= - \mathbb{E}_{q\left(\btheta\right)}\left[\log p\left(\mathcal{D} | \btheta\right)\right] + \mathbb{D}_{KL}\left[\left. q\left(\btheta\right) \right\| p\left(\btheta \right)\right]\\
	&= - \mathbb{E}_{q\left(\btheta\right)}\left[\log p\left(\mathcal{D} | \btheta\right) + \log p\left(\btheta\right)\right] -\mathcal{H}\left[q\left(\btheta\right)\right]
	\end{aligned}
\end{equation}
\begin{equation} \label{Eq:Bayes_reform}
	p\left(\btheta|\mathcal{D}\right) = \underset{q\left(\btheta\right)\in \mathcal{P}}{\arg\min} J_{B}\left(\btheta\right) 
\end{equation}
where $\mathcal{H}\left[p\right]:=-\mathbb{E}_{p}\left[\log p\right]$ denotes the entropy. Therefore, the posterior distribution computation is equivalent to the minimisation of the Bayesian objective $J_{B}$ defined by Eq. \eqref{Eq:Bayes_reform_objective} when the domain of the minimisation problem is the entire set $\mathcal{P}$ of probability distributions. According to Eq. \eqref{Eq:Bayes_reform_objective}, the Bayesian objective $J_{B}$ naturally includes the negative entropy term which encodes the maximum-entropy principle \cite{Jaynes_1982,Jaynes_1988}.

The exact Bayesian approach requires marginalisation over all possible candidate posterior distributions which is essentially a computationally demanding procedure of evaluating high-dimensional integrals and hence impossible in practice. In this regard, one can take an approximate solution approach to the posterior computation problem by restricting the set of candidate probability distributions $q\left(\btheta\right)$ to a specific subclass $\mathcal{Q}$. This is equivalent to the variational inference approach \cite{Zellner_1988} which aims to find a probability distribution out of an assumed set of candidates that maximises the evidence lower bound. Thus, the solution to the posterior distribution computation problem can be approximated as
\begin{equation} \label{Eq:Bayes_approximate}
	q^{*}\left(\btheta\right) = \underset{q\left(\btheta\right)\in \mathcal{Q}}{\arg\min} J_{B}\left(\btheta\right) \approx p\left(\btheta|\mathcal{D}\right) 
\end{equation}

\subsubsection{Control as Bayesian Learning}
Now, the Bayesian learning approach can be bridged to the control problem of our interest. This can be done by relating the mathematical representation as well as the physical meaning of the control problem defined in Eq. \eqref{Eq:ProbForm_DeterSys} with each element comprising the approximate Bayesian posterior computation problem defined by Eq. \eqref{Eq:Bayes_approximate}. 

In the context of this study, $\btheta$ stands for the random variable of the policy as decribed with the same notation in Sec. \ref{Sec:ProbForm}, and the dataset is defined to be the tuple of state-control trajectories predicted at the current $t$ for $N$ sampled control functions, i.e.,
\begin{equation} \label{Eq:dataset}
	\mathcal{D}:= \left\{ \hat{\mathbf{x}}^{\left(i\right)}\left(\tau\right), \hat{\mathbf{u}}^{\left(i\right)}\left(\tau\right) \right\}_{\tau\in t:t_{f}}^{i\in 1:N}
\end{equation}
The physical meaning of each distribution is different from those in the pure machine learning problems such as regression or classification. In optimal control problems, the cost function $C_{t:t_{f}}$ is related with the likelihood function $p\left(\btheta|\mathcal{D}\right)$ in the Bayesian learning objective. Hence, it is natural to state that the objective function $J$ needs to be evaluated for the trajectories in $\mathcal{D}$. The regulariser term $R$ in the optimal control objective $J$ can be related with the parameter prior distribution in the reformulated Bayesian learning objective. Note that, in terms of online learning where each decision incurs a loss value, the objective function becomes the loss function associated with the decision made. Lastly, the policy, i.e., the control distribution, $p_{u}\left(\btheta|\left[t,t_{f}\right]\right)$ that should be optimised in the optimal control problem can be viewed as the candidate posterior distribution $q\left(\btheta\right)$. 

More specifically, the bridging relations can be expressed as follows:
\begin{align}
	p\left( \mathcal{D} | \btheta \right) & \propto \exp\left(-C_{t:t_{f}}\left(\btheta\right)\right) \label{Eq:p_likelihood_cost}\\
	p\left(\btheta\right) & \propto \exp\left(-R\left(\btheta\right)\right) \label{Eq:p_prior_regulariser}\\
	q\left(\btheta\right) & \propto p_{u}\left(\btheta | \left[t,t_{f}\right]\right) \label{Eq:q_dist_policy}
\end{align}
The apparent difference between the non-Bayesian optimal control problem and the associated Bayesian-interpreted problem descriptions is the negative entropy term. The entropy-maximising term can indeed be introduced in the non-Bayesian setting as the entropic regularisation term to promote exploration of search space and to alleviate collapsing into a low-quality local minimum. However, such entropy-maximisation effect is shown to be a natural consequence of the Bayesian approach itself.

Although the last expression in Eq. \eqref{Eq:Bayes_approximate} with the presence of a separate entropy maximisation gives deep insights about the Bayesian learning perspective towards control problems, the intermediate relation written in terms of the KL divergence of the candidate posterior distribution from the prior distribution is more useful in the MPC setup. The KL divergence does not truly qualify as a distance function unlike the Wasserstein metric, but nonetheless, it has the meaning as a statiscal quantification of the discrepancy between two probability distributions. In this sense, $\mathbb{D}_{KL}\left[\left. q\left(\btheta\right) \right\| p\left(\btheta \right)\right]$ represents how much the posterior distribution differs from the given prior distribution. 

In the MPC problems, the physical meaning of the probability distributions suggests that minimisation of the KL divergence encodes the tendency to keep the updated policy $p_{u}\left(\btheta | \left[t,t_{f}\right]\right)\in\mathcal{Q}$ in a close neighbourhood of the prior policy $p\left(\btheta\right) \in \mathcal{Q}$. Here, the prior policy can represent a given fixed nominal policy or the posterior policy updated in the previous round when the Bayesian learning is performed in the sequential online learning manner. The latter is more practically meaningful in the MPC problem, since maintaining a close distance to the previous policy before optimisation can prevent abrupt change of the control input at each time. Alternatively, the sequential update structure can be understood as a warm-starting scheme, provided that the initial policy provides at least a marginal degree of performance. 

The sequential online learning representation of the reformulated Bayesian learning problem is given as follows:
\begin{align} 
	J_{B}^{j+1}\left(\btheta\right) &:= \mathbb{E}_{p_{u}}\left[C_{t:t_{f}}\left(\btheta\right)\right] + \mathbb{D}_{KL}\left[\left. p_{u}\left(\btheta\right) \right\| p_{u}^{j}\left(\btheta \right)\right] \label{Eq:Bayes_approximate_MPC_objective}\\	
	p_{u}^{j+1}\left(\btheta \right) &= \underset{p_{u}\left(\btheta\right)\in \mathcal{Q}}{\arg\min}J_{B}^{j+1}\left(\btheta\right) \label{Eq:Bayes_approximate_MPC}
\end{align}
where the optimal posterior distribution for the $j$-th round is defined as the policy planned over the time window $\left[t_{j},t_{f}\right]$ with a fixed $t_{f}$ or a receding $t_{f} = t_{j}+T$ as
\begin{equation} \label{Eq:p_posterior_j_defn}
	p_{u}^{j}\left(\btheta\right) := p_{u}\left(\btheta | \left[t_{j},t_{f}\right]\right)
\end{equation}
With this background, the physical interpretation of the regulariser is clear from the relation in Eq. \eqref{Eq:p_prior_regulariser} and the role of prior distribution being the policy updated in the previous round. The regulariser is associated with the quantification of the distance between successive optimal solutions at the level of parameter while the KL divergence term measures at the level of probability distribution in the same class $\mathcal{Q}$.

\subsubsection{Exact Optimal Solution for Control Problems}
\paragraph{Non-sequential Bayesian Learning}
Before proceeding further with the variational approximate Bayesian formulation, let us recapitulate the optimal solution to the control problem in view of the exact Bayesian formulation. The optimal policy minimising the Bayesian objective is in principle the posterior distribution as discussed in the previous section. By substituting Eqs. \eqref{Eq:p_likelihood_cost}-\eqref{Eq:q_dist_policy} into Eq. \eqref{Eq:Bayes_original}, the optimal policy obtained as the posterior distribution can be written as
\begin{equation} \label{Eq:p_posterior_full_opt}
	\begin{aligned}
	p_{u}^{*}\left(\btheta | \left[t,t_{f}\right]\right) &= p\left(\btheta | \mathcal{D}\right) =\frac{p\left(\mathcal{D} | \btheta\right) p\left(\btheta\right)}{\int p\left(\mathcal{D} | \btheta\right) p\left(\btheta\right) d\btheta}\\
	&=\frac{1}{Z} \exp\left[ -C_{t:t_{f}}\left(\btheta\right) - R\left(\btheta\right) \right]
	\end{aligned}
\end{equation}
where
\begin{equation} \label{Eq:partition_defn}
	Z := \int \exp\left[ -C_{t:t_{f}}\left(\btheta\right) - R\left(\btheta\right) \right] d\btheta 
\end{equation}
The optimal Bayesian objective can be obtained by evaluating the function $J_{B}$ defined in Eq. \eqref{Eq:Bayes_reform_objective}, with Eqs. \eqref{Eq:p_posterior_full_opt} and \eqref{Eq:J_Bayes_opt} as
\begin{equation} \label{Eq:J_Bayes_opt}
	\begin{aligned}
		J_{B}^{*} &= \mathbb{E}_{p_{u}^{*}\left(\btheta\right)}\left[C_{t:t_{f}}\left(\btheta\right)+R\left(\btheta\right)\right] + \mathbb{E}_{p_{u}^{*}\left(\btheta\right)}\left[\log p_{u}^{*}\left(\btheta\right) \right]\\
		&= \mathbb{E}_{p_{u}^{*}\left(\btheta\right)}\left[C_{t:t_{f}}\left(\btheta\right)+R\left(\btheta\right)\right] \\
		&\quad + \mathbb{E}_{p_{u}^{*}\left(\btheta\right)}\left[\log\frac{1}{Z} \exp\left[ -C_{t:t_{f}}\left(\btheta\right) - R\left(\btheta\right) \right] \right]\\
		&= - \mathbb{E}_{p_{u}^{*}\left(\btheta\right)}\left[\log Z \right] = -\log Z\\
		&= -\log \int \exp\left[ -C_{t:t_{f}}\left(\btheta\right) - R\left(\btheta\right) \right] d\btheta 
	\end{aligned}
\end{equation}
In summary, the optimal policy in Eq. \eqref{Eq:p_posterior_full_opt} turns out to involve the softmax-type operation, and the optimal Bayesian objective in Eq. \eqref{Eq:J_Bayes_opt} has the form of log-sum-exp function. 

\paragraph{Sequential Bayesian Learning}
In the sequential online learning setup, the exact Bayesian optimal solution is recursively updated by treating the result of the previous round as the prior distribution in the current round. By relating the regulariser $R\left(\btheta\right)$ with the previous optimal solution through Eq. \eqref{Eq:p_prior_regulariser}, the sequential update structure turns the expressions in Eqs. \eqref{Eq:p_posterior_full_opt}-\eqref{Eq:J_Bayes_opt} into the following recursive form:
\begin{align} 
	Z^{j+1} 
	&= \mathbb{E}_{p_{u}^{j}}\left[\exp\left[ -C_{t:t_{f}}\left(\btheta\right) \right]\right] \label{Eq:partition_seq}\\
	p_{u}^{j+1}\left( \btheta \right) &= \frac{\exp\left[ -C_{t:t_{f}}\left(\btheta\right) \right]}{\mathbb{E}_{p_{u}^{j}}\left[\exp\left[ -C_{t:t_{f}}\left(\btheta\right) \right]\right]}  p_{u}^{j}\left( \btheta \right)  \label{Eq:p_posterior_full_opt_seq}\\
	\left(J_{B}^{j+1}\right)^{*} &= -\log \mathbb{E}_{p_{u}^{j}}\left[\exp\left[ -C_{t:t_{f}}\left(\btheta\right) \right]\right] \label{Eq:J_Bayes_opt_seq}
\end{align}
where the optimal Bayesian objective for sequential online learning $J_{B}^{j+1}$ follows the definition in Eq. \eqref{Eq:Bayes_approximate_MPC_objective}.

The exact optimal solution method is computationally intractable due the necessity of evaluating the integral defined over the entire space of $\btheta$ for marginalisation. Monte-Carlo sampling-based integral approximation can be introduced even at this stage in principle before carrying out variational inference. However, recursive update of a non-parametric policy is not practical because of the excessive memory requirement. In the MPC-type control problems, the Monte-Carlo integration requires evaluation of the cost function $C_{t:t_{f}}$ for multiple trajectories predicted with different control functions $\hat{\mathbf{u}}^{\left(i\right)}\left(\tau\right)$ sampled from the policy with the fixed initial condition $\mathbf{x}_{t}$. 

\subsection{Optimisation Algorithm}
\subsubsection{Posterior Approximation with Exponential Family \cite{Nielsen_2022}}
The function class $\mathcal{Q}$ should be selected to solve the optimisation problem of the sequential online learning form given in Eq. \eqref{Eq:Bayes_approximate_MPC}. In the following developments, the BL-MPC framework will assume $\mathcal{Q}$ to be the minimal exponential family of probability distributions for the random variable $\btheta$ that can be expressed as
\begin{equation} \label{Eq:exp_fam_defn}
	q\left(\btheta\right) = \rho\left(\btheta\right)\exp\left[ \left< \bfeta, \mathbf{T}\left(\btheta\right)\right> - A\left(\bfeta\right) \right]
\end{equation}
where $\bfeta$ is the natural parameter, $\left<\cdot, \cdot\right>$ indicates the inner product, $\rho\left(\btheta\right)$ is the base/carrier measure which is also known as a scaling constant, $\mathbf{T}\left(\btheta\right)$ is the vector of sufficient statistics which are linearly independent, and $A\left(\bfeta\right):=\log \int \rho\left(\btheta\right)\exp\left[\left< \bfeta, \mathbf{T}\left(\btheta\right)\right> \right] d\btheta$ is the finite, strictly convex, and differentiable log-partition function. The exponential family can be described with another parametrisation with respect to the expectation parameter defined by $\bmu := \mathbb{E}_{q}\left[\mathbf{T}\left(\btheta\right)\right]$ which is in a bijective relationship with the natural parameter $\bfeta$. Consider the Legendre transformation of $A$ given by
\begin{equation} \label{Eq:Legendre_dual}
	A^{*}\left(\bmu\right) = \sup\limits_{\bfeta'\in\mathcal{H}} \left[ \left< \bfeta', \bmu \right> - A\left(\bfeta'\right) \right]
\end{equation}
where $\mathcal{H} = \left\{\bfeta | A\left(\bfeta\right) < \infty \right\}$. Note that the reverse mapping has the similar form
\begin{equation} \label{Eq:Legendre_dual_reverse}
	A\left(\bfeta\right) = \sup\limits_{\bmu'\in\mathcal{M}} \left[ \left< \bfeta, \bmu' \right> - A^{*}\left(\bmu'\right) \right]
\end{equation}
where $\mathcal{M} = \left\{\bmu | A^{*}\left(\bmu\right)<\infty\right\}$. The duality that exists between two parametrisations can be written as
\begin{equation} \label{Eq:nat_expec_params}
	\begin{aligned}
		\bmu 	&= \nabla_{\bfeta} A\left(\bfeta\right) \\
		\bfeta 	&= \nabla_{\bmu} A^{*}\left(\bmu\right) \\
		\nabla_{\bfeta}A &= \left(\nabla_{\bmu} A^{*}\right)^{-1}
	\end{aligned}
\end{equation}

The Bregman divergence associated with a strictly convex potential function $\psi$ is defined as
\begin{equation} \label{Eq:Bregman_div}
	\mathbb{D}_{\psi}\left[\left. \bxi_{1} \right\| \bxi_{2} \right] := \psi\left(\bxi_{1}\right) - \psi\left(\bxi_{2}\right) - \left<\bxi_{1}-\bxi_{2}, \nabla_{\xi}\psi\left(\bxi_{2}\right)\right>
\end{equation}
Following from the duality in Eq. \eqref{Eq:nat_expec_params} and the definition of Bregman divergence in Eq. \eqref{Eq:Bregman_div}, it is well-known that the KL divergence between the probability distributions $q_{1}$ and $q_{2}$ in the same exponential family $\mathcal{Q}$ and the Bregman divergences between parameters are related to each other \cite{Nielsen_2022} as
\begin{equation} \label{Eq:KL_Bregman_div}
	\mathbb{D}_{KL}\left[\left. q_{1}\left(\btheta\right) \right\| q_{2}\left(\btheta\right) \right] = \mathbb{D}_{A^{*}}\left[\left. \bmu_{1} \right\| \bmu_{2} \right] = \mathbb{D}_{A}\left[\left. \bfeta_{2} \right\| \bfeta_{1} \right] 
\end{equation}
Therefore, the alternative way of specifying the exponential family is given as
\begin{equation} \label{Eq:exp_fam_alt}
	q\left(\btheta\right) = \rho\left(\btheta\right)\exp\left[ -\mathbb{D}_{A^{*}}\left[ \left. \mathbf{T}\left(\btheta\right)\right\|\bmu\right] + A^{*}\left(\mathbf{T}\left(\btheta\right)\right) \right]
\end{equation}

The gradients of an objective function with respect to the dual parametrisations can be related to each other through Eq. \eqref{Eq:nat_expec_params}. Let $I\left(\bfeta\right) = I_{*}\left(\bmu\right)$ be two different representations in terms of each parametrisation for an identical function. By using the first line of Eq. \eqref{Eq:nat_expec_params}, we have
\begin{equation} \label{Eq:nat_expec_grads}
	\begin{aligned}
		\nabla_{\bfeta} I\left(\bfeta\right) &= \nabla_{\bfeta} \bmu \nabla_{\bmu} I_{*}\left(\bmu\right) = \nabla_{\bfeta} \nabla_{\bfeta} A\left(\bfeta\right) \nabla_{\bmu} I_{*}\left(\bmu\right)\\
		&= \nabla_{\bfeta \bfeta}^{2} A\left(\bfeta\right) \nabla_{\bmu} I_{*}\left(\bmu\right):=\mathbf{F}\left(\bfeta\right) \nabla_{\bmu} I_{*}\left(\bmu\right)
	\end{aligned}
\end{equation}
where $\mathbf{F}\left(\bfeta\right)=\mathbb{E}_{q}\left[\nabla_{\bfeta}\log q\left(\btheta\right)\nabla_{\bfeta}\log q\left(\btheta\right)^{T}\right]$ denotes the Fisher information matrix. Therefore, natural gradients are the standard gradients scaled by Fisher information matrix
\begin{equation} \label{Eq:nat_expec_grads_scaling}
	\nabla_{\bmu} = \mathbf{F}\left(\bfeta\right)^{-1} \nabla_{\bfeta}
\end{equation}

\subsubsection{Optimality Condition}
By incorporating Eq. \eqref{Eq:KL_Bregman_div}, the Bayesian objective given by Eq. \eqref{Eq:Bayes_approximate_MPC_objective} can be rewritten in terms of the Bregman divergence as
\begin{equation} \label{Eq:Bayes_approximate_MPC_Bregman}
	J_{B}^{j+1}\left(\btheta \right)= \mathbb{E}_{p_{u}}\left[C_{t:t_{f}}\left(\btheta\right)\right] + \mathbb{D}_{A^{*}}\left[\left. \bmu \right\| \bmu^{j} \right]
\end{equation}
where $\bmu$ and $\bmu^{j}$ represent the expectation parameters for $p_{u}\left(\btheta\right)$ and $p_{u}^{j}\left(\btheta\right)$ in $\mathcal{Q}$, respectively.

The gradient of the Bayesian objective $J_{B}^{j+1}$ defined in Eq. \eqref{Eq:Bayes_approximate_MPC_objective} for sequential online learning problem in the distribution space vanishes when evaluated at the optimal approximate posterior distribution. If the exponential family is selected as the class of approximate posteriors, the form of $J_{B}^{j+1}$ further specialises to the one in Eq. \eqref{Eq:Bayes_approximate_MPC_Bregman}. Using Eqs. \eqref{Eq:nat_expec_params}, \eqref{Eq:Bregman_div}, and \eqref{Eq:nat_expec_grads}, the optimality condition for sequential learning can be expressed in terms of the gradient taken with respect to natural parameter as
\begin{equation} \label{Eq:opt_cond}
	\resizebox{0.89\hsize}{!}{$\displaystyle
	\begin{aligned}
	&\nabla_{\bfeta} \left. J_{B}^{j+1}\left(\bfeta\right) \right|_{\bfeta =\bfeta^{*}} = \nabla_{\bfeta} \left( \mathbb{E}_{p_{u}}\left[C_{t:t_{f}}\left(\btheta\right)\right] + \mathbb{D}_{A^{*}}\left[\left. \bmu \right\| \bmu^{j} \right] \right)_{\bfeta =\bfeta^{*}}\\
	&= \mathbf{F}\left(\bfeta^{*}\right) \nabla_{\bmu} \left( \mathbb{E}_{p_{u}}\left[C_{t:t_{f}}\left(\btheta\right)\right] + \mathbb{D}_{A^{*}}\left[\left. \bmu \right\| \bmu^{j} \right] \right)_{\bmu =\bmu^{*}}\\
	&= \mathbf{F}\left(\bfeta^{*}\right)  \left(\nabla_{\bmu}\left.\mathbb{E}_{p_{u}}\left[C_{t:t_{f}}\left(\btheta\right)\right] \right|_{\bmu =\bmu^{*}} + \bfeta^{*}-\bfeta^{j}\right) =\mathbf{0}
	\end{aligned}
	$}
\end{equation}
where $\bmu^{*} = \nabla_{\bfeta}A\left(\bfeta^{*}\right)$. In turn, the optimal posterior approximation has its natural parameter equal to the prior natural parameter minus the gradient of the expected cost taken with respect to the expectation parameter
\begin{equation} \label{Eq:opt_nat_param}
	\bfeta^{*} = \bfeta^{j} -\nabla_{\bmu}\left.\mathbb{E}_{p_{u}}\left[C_{t:t_{f}}\left(\btheta\right)\right] \right|_{\bmu =\bmu^{*}}
\end{equation}

\subsubsection{Parameter Update Rule}
Various parameter optimisation algorithms, especially the gradient-descent-based algorithms, can be employed to minimise the Bayesian objective $J_{B}^{j+1}$ at each round of sequential update. Even the methods based on continuous evolution of gradient flow dynamics are also applicable. Among others, the natural gradient descent algorithm is particularly useful when the posterior distribution is chosen to be approximated with the exponential family.

Natural gradient descent algorithm was proposed in \cite{Amari_1998} and enhanced recently in \cite{Martens_2020} as a systematic method paying attention to the information geometry in the space of probability distributions through preconditioning with the curvature carried in Fisher information matrix. The main benefit of natural gradient descent is known to be the correct computation of the update step size and direction that substantially accelerates the rate of convergence. Motivated by the optimality condition for the sequential online learning, the natural gradient descent update rule can be written as
\begin{equation} \label{Eq:nat_grad_rule}
	\resizebox{0.89\hsize}{!}{$\displaystyle
	\begin{aligned}
		\bfeta_{k+1}^{j+1} 	&= \bfeta_{k}^{j+1} - \gamma_{k} \mathbf{F}\left(\bfeta_{k}^{j+1}\right)^{-1} \nabla_{\bfeta} J_{B}^{j+1}\left(\bfeta_{k}^{j+1}\right)\\
		&=\left(1-\gamma_{k}\right)\bfeta_{k}^{j+1}  -\gamma_{k}\left(\nabla_{\bmu}\left.\mathbb{E}_{p_{u}}\left[C_{t:t_{f}}\left(\btheta\right)\right] \right|_{\bmu_{k}^{j+1}} -\bfeta^{j}\right)
	\end{aligned}
	$}
\end{equation}
where $\gamma_{k}$ is the learning rate, $\bfeta_{k}^{j+1}$ denotes the natural parameter of the policy $p_{u}^{j+1}$ for the $(j+1)$-th round updated in the $k$-th iteration of optimisation process, and $\bfeta^{j}$ is the natural parameter optimised in the $j$-th round. Equation \eqref{Eq:nat_grad_rule} turns out to be the BLR of \cite{Khan_2021} applied to the MPC problem and it does not require evaluation of the inverse of the Fisher information matrix for implementation.

\begin{rem}
	The log objective $\tilde{J}_{B}^{j+1}:=\log J_{B}^{j+1}$ can be considered instead of $J_{B}^{j+1}$ in the recursive update following Eq. \eqref{Eq:nat_grad_rule}. The update step in this case will be normalised.
\end{rem}

\subsubsection{BL-MPC Framework with Warm-Starting Initialisation}
The BL-MPC approach provides a control profile $\bar{\mathbf{u}}^{j+1}\left(\tau\right)$ spanning the future time horizon $\left[t_{j+1}, t_{f}\right]$ at each planning round by computing the expectation of policy. To warm-start the optimisation process, the initial guess $\bfeta_{0}^{j+1}$ for the current round can be generated by forward shifting of the parameter $\bfeta^{j}$ obtained in the previous round. Algorithm \ref{Algo:BLMPC} summarises the BL-MPC algorithm for posterior approximation with exponential family including such warm-starting initialisation $\Phi$.

\begin{algorithm}[htb!]
	\caption{Bayesian Learning Model Predictive Control - Exponential Family ($\mathsf{BL}$-$\mathsf{MPC}$-$\mathsf{Exp}$)}	\label{Algo:BLMPC}
	\begin{algorithmic}
		\renewcommand{\algorithmicrequire}{\textbf{Input:}}
		\REQUIRE $t_{j+1}$, $t_{f}$, $\mathbf{x}_{t}$, $\hat{\mathbf{f}}$, $C$, $p_{u}^{j+1}$, $\bfeta^{j}$, $\gamma_{k}$
		\renewcommand{\algorithmicensure}{\textbf{Output:}}
		\ENSURE $\bfeta^{j+1}$, $\bar{\mathbf{u}}^{j+1}\left(\tau\right)$ for $\tau \in \left[t_{j+1},t_{f}\right]$
		
		\STATE $k \leftarrow 0$ and $\bfeta_{0}^{j+1} \leftarrow \Phi\left(\bfeta^{j}\right)$

		\WHILE{$\bfeta_{k}^{j+1}$ is not converged} 
			\STATE \textsc{Sample} $\hat{\mathbf{u}}^{\left(i\right)}\left(\tau\right) \sim p_{u}^{j+1}\left(\left.\btheta \right| \bmu_{k}^{j+1}\right)$  for $i=1,\cdots,N$
			\STATE \textsc{Simulate} $\hat{\mathbf{x}}^{\left(i\right)}\left(\tau\right) = \mathbf{x}_{t} + \int_{t}^{\tau} \hat{\mathbf{f}}\left(\xi, \hat{\mathbf{x}}^{\left(i\right)}\left(\xi\right), \hat{\mathbf{u}}^{\left(i\right)}\left(\xi\right) \right)d\xi$
			\STATE \textsc{Evaluate} $\mathbf{g}_{k}^{j+1} \leftarrow \nabla_{\bmu}\left.\mathbb{E}_{p_{u}}\left[C_{t:t_{f}}\left(\btheta\right)\right] \right|_{\bmu_{k}^{j+1}}$
			\STATE $\bfeta_{k+1}^{j+1} \leftarrow \left(1-\gamma_{k}\right)\bfeta_{k}^{j+1} -\gamma_{k}\left(\mathbf{g}_{k}^{j+1}-\bfeta^{j}\right)$
			\STATE $k \leftarrow k+1$
		\ENDWHILE
		\RETURN $\bfeta^{j+1} \leftarrow \bfeta_{k}^{j+1}$ and $\bar{\mathbf{u}}^{j+1}\left(\tau\right) \leftarrow \mathbb{E}_{p_{u}^{j+1}}\left[\btheta\right]$
	\end{algorithmic}
\end{algorithm}

\begin{rem}
	A mirror descent step in the space of expectation parameter $\bmu$ is equivalent to a natural gradient descent step in the space of natural parameter $\bfeta$. In this way, the BL-MPC employing natural gradient descent as the optimisation rule is certainly connected to the DMD-MPC framework. Nevertheless, the philosophy of the BL-MPC differs from that of the DMD-MPC. The variational approximate Bayesian approach interprets the Bregman divergence term in Eq. \eqref{Eq:Bayes_approximate_MPC_Bregman} as the object governed by the chosen posterior approximation rather than as an independent element allowed for user definition. This is a notable difference from the DMD-MPC perspective where the choice of the Bregman divergence term is discussed as a tuning knob for generating different MPC algorithms but without fully appreciating the connection to variational Bayesian formulation. Furthermore, the BL-MPC framework considers trajectory prediction by solving ODE dynamics with sampling control functions from a distribution over function, whereas the DMD-MPC considers discrete-time state transition with control sequences being pointwise-independent are sampled from distribution over parameter. Lastly, assuming that the computation can be performed very rapidly, the BL-MPC does not limit the number of iterations for optimisation in each round only to one.
\end{rem}

\section{Gaussian Example} \label{Sec:GaussExample}
This section presents a concrete example of the BL-MPC approach for posterior approximation class $\mathcal{Q}$ given by the Gaussian family $p_{u}\left(\btheta\right) = \mathcal{N}\left(\btheta; \mathbf{m}, \mathbf{\Sigma}\right)$. The natural and expectation parameters of a Gaussian distribution has two components.
\begin{equation} \label{Eq:nat_expec_params_Gauss}
	\begin{aligned}
		\bfeta^{\left(1\right)} &= \mathbf{\Sigma}^{-1}\mathbf{m} \quad	&	\bmu^{\left(1\right)} &= \mathbb{E}_{p_{u}}\left[\btheta\right]=\mathbf{m}\\
		\bfeta^{\left(2\right)} &= -\frac{1}{2}\mathbf{\Sigma}^{-1}	\quad & \bmu^{\left(2\right)} &=\mathbb{E}_{p_{u}}\left[\btheta\btheta^{T}\right]= \mathbf{\Sigma} + \mathbf{m}\mathbf{m}^{T}
	\end{aligned}
\end{equation}
The gradients of the expected cost with respect to the expectation parameters are required to implement the update rule of Eq. \eqref{Eq:nat_grad_rule}. Applying the chain rule yields
\begin{equation} \label{Eq:grad_expec_cost_chain}
	\begin{aligned}
		\nabla_{\bmu^{\left(1\right)}} \mathcal{E}_{k} &= \nabla_{\mathbf{m}} \mathcal{E}_{k} - 2\left[\nabla_{\mathbf{\Sigma}}\mathcal{E}_{k}\right]\mathbf{m}\\
		\nabla_{\bmu^{\left(2\right)}} \mathcal{E}_{k} &= \nabla_{\mathbf{\Sigma}}\mathcal{E}_{k} 
	\end{aligned}
\end{equation}
where $\mathcal{E}_{k} := \left.\mathbb{E}_{p_{u}}\left[C_{t:t_{f}}\left(\btheta\right)\right] \right|_{\bmu_{k}^{j+1}}$. Let the prior be $p_{u}^{j}\left(\btheta\right)= \mathcal{N}\left(\btheta;\mathbf{m}^{j}, \mathbf{\Sigma}^{j}\right)$. Then, by substituting Eqs. \eqref{Eq:nat_expec_params_Gauss}-\eqref{Eq:grad_expec_cost_chain} into Eq. \eqref{Eq:nat_grad_rule}, we have
\begin{equation} \label{Eq:nat_grad_rule_Gauss}
	\resizebox{0.89\hsize}{!}{$\displaystyle
	\begin{aligned}
		\left(\mathbf{\Sigma}_{k+1}^{j+1}\right)^{-1} &= \left(1-\gamma_{k}\right)\left(\mathbf{\Sigma}_{k}^{j+1}\right)^{-1} + \gamma_{k} \left[2\nabla_{\mathbf{\Sigma}}\mathcal{E}_{k}+\left(\mathbf{\Sigma}^{j}\right)^{-1}\right]\\
		\mathbf{m}_{k+1}^{j+1} &= \mathbf{m}_{k}^{j+1} - \gamma_{k}\mathbf{\Sigma}_{k+1}^{j+1}\left[\nabla_{\mathbf{m}}\mathcal{E}_{k} + \left(\mathbf{\Sigma}^{j}\right)^{-1}\left(\mathbf{m}_{k}^{j+1}-\mathbf{m}^{j}\right)\right]
	\end{aligned}
	$}
\end{equation}
Theorems of Bonnet and Price \cite{Opper_2009} give
\begin{equation} \label{Eq:Bonnet_Price}
	\begin{aligned}
		\nabla_{\mathbf{m}} \mathcal{E}_{k} &= \left.\mathbb{E}_{p_{u}}\left[\nabla_{\btheta}C_{t:t_{f}}\left(\btheta\right)\right] \right|_{\bmu_{k}^{j+1}}\\
		\nabla_{\mathbf{\Sigma}} \mathcal{E}_{k} &= \frac{1}{2} \left.\mathbb{E}_{p_{u}}\left[\nabla_{\btheta}^{2}C_{t:t_{f}}\left(\btheta\right)\right] \right|_{\bmu_{k}^{j+1}}
	\end{aligned}
\end{equation}
Monte-Carlo sampling can be used in practice for estimation of the gradient and the Hessian along with Gauss-Newton approximation. Suppose that $N$ samples $\btheta^{\left(i\right)} \sim p_{u_{k}}^{j+1}\left(\btheta\right)=\mathcal{N}\left(\btheta;\mathbf{m}_{k}^{j+1},\mathbf{\Sigma}_{k}^{j+1}\right)$ form the dataset $\mathcal{D}$ consisting of trajectory predictions $\left(\hat{\mathbf{x}}^{\left(i\right)}, \hat{\mathbf{u}}^{\left(i\right)} \right)$. Then, the quantities in Eq. \eqref{Eq:Bonnet_Price} can be estimated empirically as
\begin{equation} \label{Eq:MC_grad_exp_cost}
	\begin{aligned}
		\left.\mathbb{E}_{p_{u}}\left[\nabla_{\btheta}C_{t:t_{f}}\left(\btheta\right)\right] \right|_{\bmu_{k}^{j+1}} &\approx \frac{1}{N}\sum_{i=1}^{N}\mathbf{g}_{k}^{j+1,\left(i\right)}\\
		\left.\mathbb{E}_{p_{u}}\left[\nabla_{\btheta}^{2}C_{t:t_{f}}\left(\btheta\right)\right] \right|_{\bmu_{k}^{j+1}} &\approx \frac{1}{N}\sum_{i=1}^{N}\mathbf{g}_{k}^{j+1,\left(i\right)}\left(\mathbf{g}_{k}^{j+1,\left(i\right)}\right)^{T}
	\end{aligned}
\end{equation}
where $\mathbf{g}_{k}^{j+1,\left(i\right)} := \nabla_{\btheta} C_{t:t_{f}}\left(\btheta^{\left(i\right)}\right)$ is the per-sample gradient. 

\section{Conclusion} \label{Sec:Concls}
This study developed a Bayesian learning perspective towards model predictive control (MPC) based on sampling of control functions. The optimal control problem description was connected to the formulation of a variational approximate Bayesian learning problem, based on the notion of sequential online learning in particular. The Bayesian learning MPC approach was established as a general solution framework governed by the choice of the class of posterior approximation. The special case of exponential family recovered and complemented the existing knowledge on Bayesian learning rule and dynamic mirror descent MPC. Regarding practical applications, the Bayesian learning MPC framework has its strengths as an online correction algorithm that performs sequential update of an initial baseline policy provided by non-Bayesian offline training of neural networks or optimisation of nominal trajectories. Contrary to the traditional near-optimal control methods, the proposed algorithm can solve for optimal online action considering general nonlinear dynamics and non-smooth non-quadratic objective functions.


\bibliographystyle{IEEEtran}
\bibliography{BL_MPC}
\end{document}